\def\year{2020}\relax
\documentclass[letterpaper]{article} 
\usepackage{aaai20}  
\usepackage{times}  
\usepackage{helvet} 
\usepackage{courier}  
\usepackage[hyphens]{url}  
\usepackage{graphicx} 
\usepackage{graphicx}  
\usepackage{amsmath}
\usepackage{amsfonts}
\usepackage{bm}
\usepackage{multirow}
\usepackage{booktabs}
\usepackage[figuresright]{rotating}
\usepackage{subcaption}
\usepackage{array}

\title{Crowdfunding Dynamics Tracking: A Reinforcement Learning Approach}

\author{Jun Wang\textsuperscript{\rm 1}, Hefu Zhang\textsuperscript{\rm 1},  Qi Liu\textsuperscript{\rm 1}\thanks{Corresponding Author.}, Zhen Pan\textsuperscript{\rm 1}, Hanqing Tao\textsuperscript{\rm 1}\\
	\textsuperscript{\rm 1}Anhui Province Key Laboratory of Big Data Analysis and Application,\\ School of Computer Science and Technology, University of Science and Technology of China \\
	\{wjhxsg, zhf2011\}@mail.ustc.edu.cn, qiliuql@ustc.edu.cn, \{pzhen, hqtao\}@mail.ustc.edu.cn,
}
 
\begin{document}

\maketitle

\begin{abstract}
Recent years have witnessed the increasing interests in research of crowdfunding mechanism. In this area, dynamics tracking is a significant issue but is still under exploration. Existing studies either fit the fluctuations of time-series or employ regularization terms to constrain learned tendencies. However, few of them take into account the inherent decision-making process between investors and crowdfunding dynamics. To address the problem, in this paper, we propose a Trajectory-based Continuous Control for Crowdfunding (TC3) algorithm to predict the funding progress in crowdfunding. Specifically, actor-critic frameworks are employed to model the relationship between investors and campaigns, where all of the investors are viewed as an agent that could interact with the environment derived from the real dynamics of campaigns. Then, to further explore the in-depth implications of patterns (i.e., typical characters) in funding series, we propose to subdivide them into \textit{fast-growing} and \textit{slow-growing} ones. Moreover, for the purpose of switching from different kinds of patterns, the actor component of TC3 is extended with a structure of options, which comes to the TC3-Options. Finally, extensive experiments on the Indiegogo dataset not only demonstrate the effectiveness of our methods, but also validate our assumption that the entire pattern learned by TC3-Options is indeed the U-shaped one.
\end{abstract}

\section{Introduction}\label{sec:intro}
In recent years, crowdfunding has rapidly developed into a popular way of financial investment. It is an emerging approach that aims to solicit funds from individuals rather than traditional venture investors, such as angel investors and banks. More and more people are willing to launch a project (or named \textit{campaign} on the Internet) for different purposes. Indeed, tremendous efforts have been made by researchers to comprehend the internal mechanism in crowdfunding.
\begin{figure}[t]
	\centering
	\includegraphics[width=0.98\columnwidth]{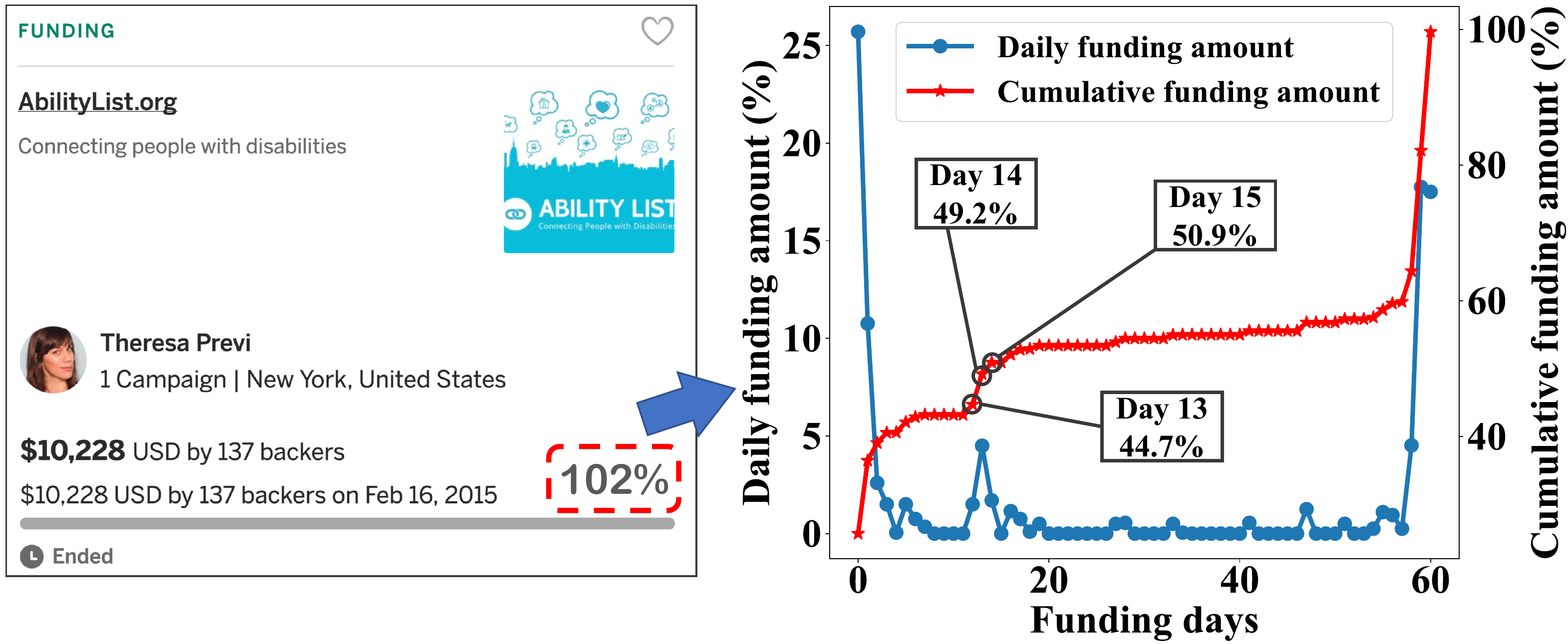}
	\caption{Funding series of a campaign example.}
	\label{fig:example}
\end{figure}

Most of the existing works focus on analyzing factors that affect the final results and predicting the probability of success. However, dynamics tracking, i.e., predicting the funding progress of campaigns, is still a problem under research. As shown in Figure~\ref{fig:example}, the funding progress of a campaign means the cumulative funding amount expressed as a percentage concerning the pledged goal (e.g, this ended campaign reached 102\% of the goal). The task concentrates on forecasting a series of percentages for campaigns that are still funding (e.g, in the 12th day, the future progress series is 44.7\%, 49.2\%, 50.9\%, ...). Actually, it is a meaningful question for funding raisers and investors. For raisers, they may acquire the forthcoming expectations of campaigns and could make quick adaptations to the market. While for potential backers, before making their possible backing decisions, they will get more detailed advice for estimating the following tendencies. Some methods have been explored in the literature, including hierarchical regression~\cite{zhao2017tracking} and basis-synthesis techniques~\cite{ren2018tracking}. Moreover, others turn to predict the backing distribution of campaigns through a Seq2Seq framework~\cite{jin2019estimating}.
However, there are still challenges on funding process modeling and series pattern utilization.

On the one hand, few of these works treat the dynamic tracking as a decision-making process. Actually, the transformation of campaigns and decision process of investors affect and depend on each other, which evolves to a complicated system~\cite{xie2019success}. Hence, compared with viewing dynamics of funding series as a whole process, the inner relationship between investors and campaigns that leads to the exterior results might not be ignored. For instance, the future tendencies of campaigns (e.g., expectation of success) will impact on the investment decision of investors, and vice versa. Nevertheless,while modeling the strong relationship between investors and campaigns, it is difficult to reflect how previous backers' behaviors affect latecomers' choices, along with how backers make decisions based on prior contribution performance and future estimates.

On the other hand, it is also significant to combine series patterns while tracking the dynamics in crowdfunding. Though pattern-decomposition techniques have been investigated, in-depth implications of patterns are not reported. Indeed, the entire pattern of funding series has been examined by~\citeauthor{kuppuswamy2017crowdfunding}~\shortcite{kuppuswamy2017crowdfunding}, i.e., U-shaped pattern. Figure~\ref{fig:example} shows a typical example for explaining what exactly the U-shaped pattern means. In other words, more contributions are likely to occur at the very beginning and ending of the funding period, as compared to the middle time. The sharp increases in the initial stage are partly because of raisers' social effect and partly due to the irrational investment behaviors which can be explained by the ``Herd Effect". While the rises in the last phase are caused by ``Goal Gradient Effect"~\cite{kuppuswamy2017crowdfunding}. Hence, to precisely utilize the entire U-shaped pattern, automatically switching mechanism is required to change between sub-patterns in pace with different periods of funding cycles.


To tackle the challenges above, we first propose a model named Trajectory-based Continuous Control for Crowdfunding (TC3). Specifically, we adopt a Markov decision process (MDP) to describe the interactions between investors and campaigns. To clearly indicate all factors that influence the decisions of investors, reinforcement learning methods, especially actor-critic frameworks are employed. In our approach, the inner transformation of campaigns is regarded as an \textit{environment}. While the \textit{agent}, which interacts with the environment, is the union of investors, along with a sub-component \textit{critic} to estimate future expectations. Secondly, to explicitly discriminate different sub-patterns in the entire U-shaped pattern, we propose to subdivide the entire pattern into \textit{fast-growing} and \textit{slow-growing} parts. Then, inspired by the idea in the hierarchical reinforcement learning area that segmenting the states and generating corresponding sub-policies, we propose TC3-Options to predict the funding progress of campaigns. With the help of a options structure, TC3-Options provides the capability of switching between different sub-patterns automatically, which means the typical U-shaped pattern behind the funding series could be precisely utilized. Finally, we conduct extensive experiments on a real-world dataset. The experimental results clearly validate that our method can predict more accurately than other state-of-the-art methods and can properly select sub-policies according to different sub-patterns.


\section{Related Work}
The related works of our study can be divided into two categories: crowdfunding and reinforcement learning.

\noindent \textbf{Crowdfunding.} With the growing popularity of crowdfunding, scholars have done much research and analysis from different perspectives~\cite{zhao2019voice,liu2017enhancing,zhao2017sequential,zhang2019personalized}. Most of the previous works could be grouped into three categories: analyzing the influential factors~\cite{burtch2013empirical,kuppuswamy2017crowdfunding,mollick2014dynamics,hoegen2018investors}, predicting the funding results (i.e., success of failure)~\cite{li2016project,lee2018content,yu2018prediction,zhang2019interactive,kaminski2019predicting} and tracking the funding dynamics~\cite{zhao2017tracking,ren2018tracking}, etc. Among qualitative factors, what should be mentioned is that some scholars are committed to exploring the social effects in crowdfunding, especially the ``Herd Effect" and the ``Goal Gradient Effect"~\cite{shen2010follow,herzenstein2011strategic,kuppuswamy2017crowdfunding}, which uncovers a typical and significant pattern in funding series, i.e., U-shaped pattern. For the success rate prediction task, the accuracy can be improved by combining deep learning (DL), natural language processing (NLP) and transfer learning (TL) techniques. However, simply predicting final outcomes can not reveal the detailed process in the rest of the funding cycles. When it comes to dynamics tracking, \citeauthor{zhao2017tracking}~\shortcite{zhao2017tracking} employs a hierarchical regression model that could predict funding amounts in both campaign-level and perk-level, while other researchers adopt Fourier transformation to capture various patterns hidden behind the funding series~\cite{ren2018tracking}. However, it seems that none of these works consider the inner decision-making process between investors and campaigns, which leads to exterior funding results.


\noindent \textbf{Reinforcement Learning.} Developed from Markov decision processes (MDP)~\cite{sutton2018reinforcement},  deep reinforcement learning (DRL) has been proved to be a huge success in many domains, such as games~\cite{mnih2015human,hessel2018rainbow}, robotics~\cite{kober2013reinforcement,haarnoja2018soft} and recommender systems~\cite{chen2019top,liu2019exploiting}. Existing methods could be divided into two categories: value-based methods, where policies are indirectly acquired according to the estimated value function, and policy-based methods, where policies are directly parameterized~\cite{sutton2000policy}. Gradually, actor-critic (AC) frameworks that incorporate policy gradient methods with value estimation techniques have become a mainstream~\cite{degris2012model}. 
Among AC methods, \citeauthor{lillicrap2015continuous}~\shortcite{lillicrap2015continuous} proposed Deep Deterministic Policy Gradient (DDPG) algorithm, which is more effective when it comes to continuous action space. 
While in the area of hierarchical reinforcement learning, options structure is a popular framework for temporal abstraction~\cite{sutton1999between}. In this framework, state, action and policy seem to respectively have a hierarchical structure from different views. Moreover, the option-critic architecture was proposed under the actor-critic frameworks~\cite{bacon2017option}.

Although reinforcement learning technique is suitable for circumstances under which previous outputs affect the following inputs, leading to complex changes in series, it could hardly be directly applied to track funding dynamics due to the following two reasons. First, primitive objective functions of reinforcement learning merely pay attention to maximize future rewards, while the prediction of history still needs to be considered when forecasting funding progress. Secondly, intra-option policy gradient theorem should be adapted for the deterministic case.

\begin{figure*} [t]
	\centering
	\includegraphics[width=1.98\columnwidth]{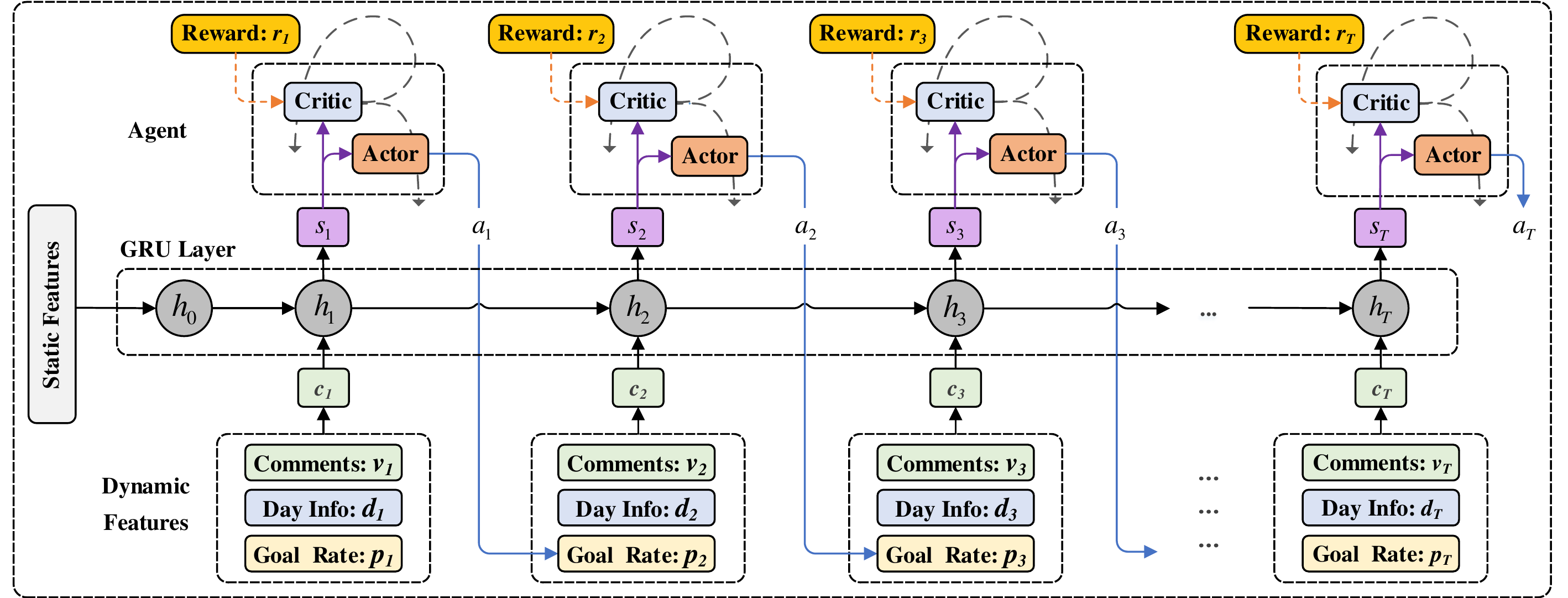}
	\caption{The framework of our TC3 and TC3-Options.}
	\label{fig:model}
\end{figure*}

\section{TC3 and TC3-Options} 
In this section, we first formally introduce the research problem, followed by the overview of basic TC3 model and final TC3-Options model. Then, we introduce the technical details in both of the models.

\begin{table}[h]
	\small
	\centering
	\caption{Main mathematical notations.}
	\begin{tabular}{c|c}
		\toprule
		Symbol     & Description \\
		\midrule
		\midrule
		$\bm{X}^i$     & static features of campaign $i$ \\
		$c^i_t$     & dynamic features of campaign $i$ in day $t$ \\
		$p^i_t$     & true funding progress of campaign $i$ in day $t$ \\
		 $\hat{p}^i_t$    & estimated funding progress of campaign $i$ in day $t$ \\
		$s_t$     & state from environment in day $t$ \\
		$a_t$     & action from actor in day $t$\\
		$r_t$     & reward from environment in day $t$ \\
		$\omega_t$     & option chosen by actor in day $t$ \\
		$\mu(s)$    & deterministic policy that chooses actions \\
		$Q(s,a)$    & function that evaluates the action $a$ in state $s$ \\
		$\pi(\omega|s)$     & stochastic policy that chooses options \\
		$\beta(s,\omega)$     & termination probability in state $s$ and option $\omega$ \\
		\bottomrule
	\end{tabular}%
	\label{tab:addlabel}%
\end{table}%

\subsection{Problem Statement}
First, we assume the process of decision-making in crowdfunding as follows. Before an investor determines whether she would contribute or not, she is likely to watch a detailed description of the campaign, including the whole story. Along with static information, some changeable information such as current funding progress, number of backers, all the updates and comments are also visible. Furthermore, the estimate of future trend is also a crucial factor that deserved to be taken into account. Finally, if the investor makes up her mind to support the campaign, she could select one perk, of which funds needed and return gained vary.

Specifically, campaign $i$ can be represented by a tuple $(\bm{X}^i,\bm{C}^i, \bm{P}^i)$. Precisely, $\bm{X}^i$ denotes static features which consist of basic information of a campaign, i.e., campaign description, perk information, a pledged goal, etc. $\bm{C}^i$ and $\bm{P}^i$ stand for dynamic features and cumulative funding progress respectively. They are both sequential data. For example, $\bm{P}^i_{1:T}=p_1^i,p_2^i,...,p_T^i$, where $T$ is the funding duration that the campaign pledges. Given the previous trajectory of campaign $i$ (i.e., $\bm{C}^i_{1:T-\tau}=c_1^i,c_2^i,...,c_{T-\tau}^i$ and $\bm{P}^i_{1:T-\tau}=p_1^i,p_2^i,...,p_{T-\tau}^i$), the goal is to predict the series of funding progress in the following $\tau$ (e.g., $\tau=5$) days (i.e., $\bm{P}^i_{T-\tau+1:T}=p_{T-\tau+1}^i,p_{T-\tau+2}^i,...,p_{T}^i$). Here, $p_t^i$ is a percentage between 0 and 1. Moreover, the dynamic features of campaign $i$ (i.e., $c_t^i$) in the $t$-th day are composed of a comments vector $v_t^i$ and a day information vector $d_t^i$.

\subsection{An Overview of TC3 and TC3-Options}
The overview of our basic TC3 is shown in Figure~\ref{fig:model}. After modeling the problem with a MDP, our approach could be generally viewed as two parts, namely an environment and an agent, along with reward signals to measure the prediction results. Specifically, the environment applies a GRU layer to integrate heterogeneous feature from campaigns while the agent includes the components of an actor and a critic. The predictions of funding progress are the outputs of the actor. While the critic is able to estimate future trends that could instruct the actor and improves the accuracy of estimates through the reward signals. Finally, we propose the TC3-Options to capture the U-shaped pattern, in which the actor is specially designed with a structure of options.

\subsection{MDP Formulation}
To particularly model the influence between behaviors of investors and dynamics of campaigns, we regard the whole of the former as one agent while thinking of the latter as an environment that can be changed by the agent. Then we apply single-agent reinforcement learning techniques to let them interact with each other. In particular, the environment is simulated from true transformation of campaigns and could be partly unchangeable (e.g., comments) and partly variable (e.g., funding progress).Therefore, we could define \textit{reward} function from the errors between true and estimated dynamics. Specifically, we model the problem described above as a MDP which comprises:  a \textit{state space} $\mathcal{S}$, an \textit{action space} $\mathcal{A}$ and a \textit{reward function} $\mathcal{R}$. Without defining \textit{state transition distribution} $\mathcal{P}$, we adopt model-free methods. In addition, a \textit{policy} $\mu$ is directly applied to select actions according to states, which is also our learning goal. Formally, we define the \textit{state}, \textit{action}, \textit{reward} in this problem as follows.

\noindent \textbf{State}. Here, we use a Gated Recurrent Unit (GRU)~\cite{chung2014empirical} layer to capture the information of dynamic inputs (i.e., day information and comments). We denote the hidden states represented by GRU layer as $h_t$, which is also the defined state of the environment, i.e., $s_t=h_t$. An extra explanation is needed that only dynamic features of the current day would be inputted and the GRU layer would aggregate useful information since the first day of the campaign, which might not contradict with the Markov assumption in the decision-making process of investors. 

\noindent \textbf{Action}. The possible percentages of the pledged goals make up the action space, which is a continuous one. Due to the unbalanced popularity of campaigns, some may rise to hundreds of times of the goals while some only achieve less than one percent. Applying deterministic policy, the output of the actor component directly means the estimated funding progress in the next day, i.e., $a_t=\hat{p}_{t+1}$. Then, to learn from quite various results that come from a series of changes, we replace the true funding progress $p_{t+1}$ in the dynamic features of the next day as the estimated one.

\noindent \textbf{Reward}. After observing state $s_t$ and taking action $a_t$, immediate reward $r_t$ with respect to $(s_t,a_t)$ needs to be returned, for measuring the error between the selected action (i.e., estimated funding progress) and the optimal action (i.e., true funding progress) in the current day. The primitive goal of reinforcement learning is to maximize discounted return $G_t$ from the current $t$-th day to the end of funding cycles, i.e., $G_t=\sum_{\tau=t}^{N}\gamma^{\tau-t}r_\tau$, where $\gamma$ is the discounted factor. Hence, we select a positive, continuous and differentiable function that decreases monotonically as the absolute error increases. In addition, to avoid violent fluctuation, the moving average technique might be required.

\subsection{Components of TC3 and TC3-Options}
In this subsection, we will first introduce the loss functions of the actor and critic components in the basic TC3. They are denoted by $L_{actor}$ and $L_{critic}$ respectively. Furthermore, the actor is extended with a structure of options and Intra-option deterministic policy gradient is derived in that sense.

\subsubsection{Basic TC3.}
Here, we derive how the actor predicts the funding progress based on the loss functions with respect to future estimates (i.e., $L_{fu}$) and past experiences (i.e., $L_{pa}$).

Actually, the actor and critic component are both parametric neural networks. While the learning goal of our models is exactly the policy, namely the function that approximated by the actor, which directly maps hidden state space to action space, i.e., $a_t=\mu_{\theta}(s_t)$ where $\theta$ denotes the parameters in the actor component. Meanwhile, the critic component evaluates the policy with respect to state-action pairs, denoted by $Q(s,a)$. Equally, it learns to estimate the future expectations of accumulated rewards, which measures the errors between possible and true future transformation after taking current action. In the $t$-th day, while the agent receives the immediate reward $r(s_t,a_t)$, it could be updated by minimizing the mean square of one-step temporal differences $\delta_t$, as shown in the following equations:
\begin{align}
	\begin{split}
	\delta_t=r_t+ \gamma Q&(s_{t+1},a_{t+1})-Q(s_t,a_t), \\
	L_{critic}&=E_{s_t\sim\rho^\mu}\delta_t^2.
	\end{split}
	\label{eq:critic}
\end{align}
On the other hand, taking advantage of estimated values from the critic, the actor is partly aimed at taking actions that could maximize discounted return $G_t=\sum_{\tau=t}^{N}\gamma^{\tau-t}r_\tau$, which equally means selecting actions that can minimize the $L_{fu}$ after the $t$-th day, where
\begin{equation}
	L_{fu}(\mu_\theta)=-E_{s_t\sim\rho^\mu}[\sum_{\tau=t}^{N}\gamma^{\tau-t}r(s_\tau,\mu_\theta(s_\tau))].
\end{equation}
Here, $\rho^\mu(s_t)$ means the distribution of state $s_t$ under the policy $\mu$ and there is no need to compute the gradient of discounted return with respect to this state distribution~\cite{silver2014deterministic}. Due to the reward $r_\tau$ is determined by the action $a_\tau$ and $a_\tau=\mu_\theta(s_\tau)$, the $L_{fu}$ is finally with respect to the policy $\mu$, of which the parameters are denoted by $\theta$.

While according to \textit{Deterministic Policy Gradient Theorem}~\cite{silver2014deterministic}, the actor can be improved in the direction of the gradient of the critic, i.e.,
\begin{equation}
	\nabla_{\theta} L_{fu}=-E_{s_t\sim\rho^\mu}[\nabla_\theta\mu_\theta(s_t)\nabla_aQ^\mu(s_t,a_t)|_{a_t=\mu_\theta(s_t)}].
	\label{eq:actor}
\end{equation}
As shown in Equation \ref{eq:actor}, the gradient of $L_{fu}$ is the negative expectation of  the product of two gradients. While the former is the gradient of policy $\mu(s_t)$ with respect to $\theta$, the latter is the gradient of value function $Q^\mu(s_t,a_t)$ with respect to action $a_t$, where $a_t$  is determined by $\mu$. Here, all expectations are actually realized by Monte-Carlo sampling.

Additionally, the actor should not only estimate the future influence but also predict the funding progress based on experienced real trajectories, namely minimizing the mean square errors between actual funding progress $p_\tau$ and estimated one $\hat{p}_\tau$ before the $t$-th day. Considering the relationship between $a_\tau$ and $\hat{p}_\tau$ that $a_\tau=\hat{p}_{\tau+1}$, this part of loss function $L_{pa}$ can be written as:
\begin{equation}
	L_{pa}=\frac{1}{t-1} \sum^t_{\tau=2}(\hat{p}_\tau-p_\tau)^2=\frac{1}{t-1} \sum^t_{\tau=2}(a_{\tau-1}-p_\tau)^2.
\end{equation}

From a practical viewpoint, it could also be regarded as a form of regularization because we want the agent to simulate future changes as closely as possible to the real ones. In other words, the actor should be updated in the direction with the maximum likelihood, especially in the initial stage where the gradient direction of the critic is uncertain.

\begin{figure} [t]
	\centering
	\includegraphics[width=0.98\columnwidth]{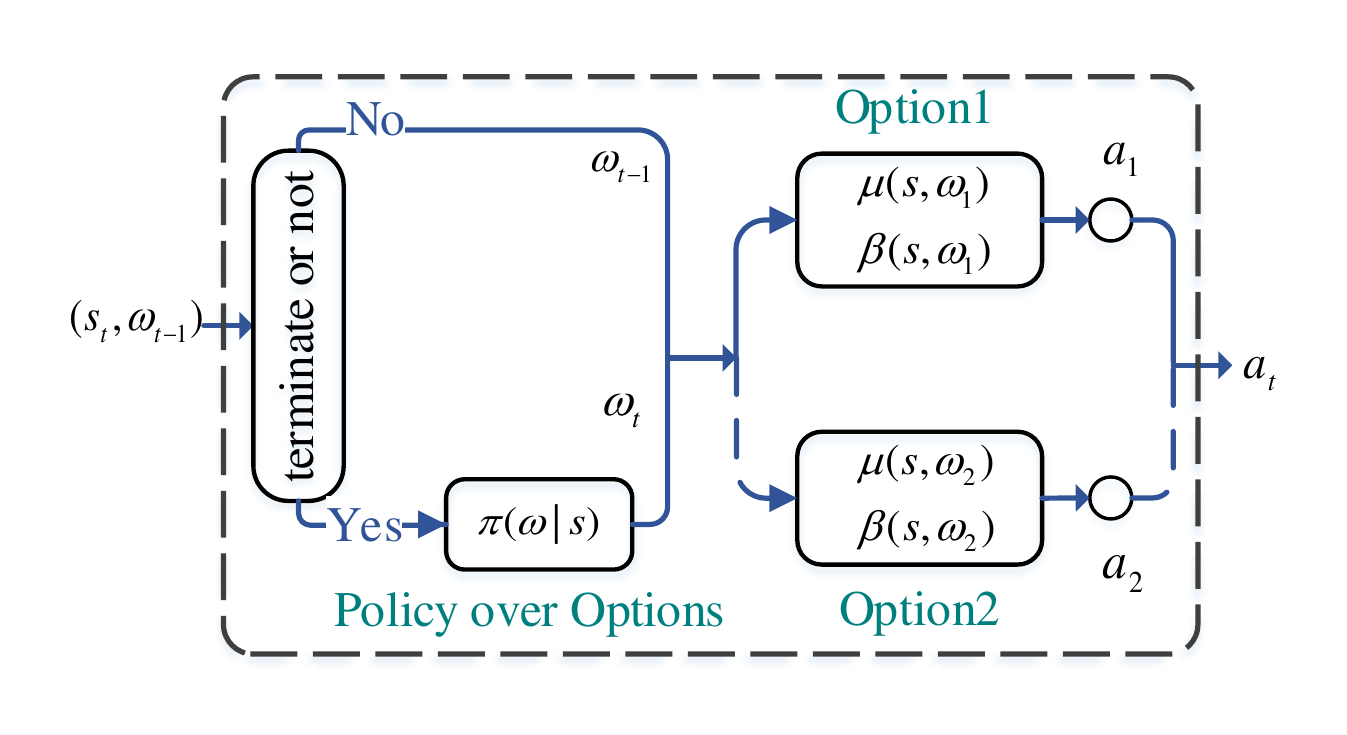}
	\caption{Actor with a structure of options.}
	\label{fig:actor}
\end{figure}

\subsubsection{Actor with Options.}
Here, we introduce how the actor component in TC3 is extended with a structure of options. After adapting previous loss functions, a defined termination loss function $L_{term}$ is added to the final $L_{actor}$.

In the beginning, we provide an informal intuition on utilizing the entire U-shaped pattern under the framework of options. After judging the stage where a campaign is in, the actor will take a sub-policy to capture those fast-growing sub-patterns (i.e., gain an optimistic prediction) if the campaign is in the beginning period or the ending period but close to the goals. On the contrary, it will switch to another sub-policy to capture slow-growing sub-patterns (i.e., obtain a smoother result) if the campaign is in the phase where the increase is gentle. To that end, primitive policy $\mu$ defined in the previous subsection is diversified with separate parameters and a high-level policy $\pi$ is needed to select proper $\mu$. 

Formally, an option $\omega$ could be represented by a tuple $(I_\omega,\mu_\omega,\beta_\omega)$. Specifically, $I_\omega$, $\mu_\omega$ and $\beta_\omega$ denote initial states, low-level policy and termination function of option $\omega$ respectively. The set of initial states with respect to option $\omega$ is a subset of the state space. In this work, we follow the assumption that all states are available to every option~\cite{bacon2017option}. The primitive policy $\mu_\omega(s)$ is the deterministic low-level policy, opposite to the stochastic high-level policy $\pi(\omega|s)$. The termination function decides the probability whether the agent will quit the current option. 

As shown in Figure \ref{fig:actor}, in the $t$-th day, the observation is the state $s_t$ and the option of last day $\omega_{t-1}$. If terminating, the agent would select a new option $\omega_t$ according to stochastic high-level policy $\pi(\omega|s_t)$, otherwise $\omega_t=\omega_{t-1}$. Then, the action to be taken is determined by $\mu_{\omega_t}(s_t)$. With next state $s_{t+1}$ received, the agent terminates this option in the next day with the probability of $\beta_{\omega_t}(s_{t+1})$.

Under specific circumstances where low-level policy $\mu$ is stochastic, \textit{Intra-Option Policy Gradient Theorem} has been derived~\cite{bacon2017option}. However, we employ a deterministic one here. The corresponding loss functions should be modified. The basic idea behind the following equations is that state-option pair $(s,\omega)$ now performs as an extension of the primitive state. Hence, $Q(s,a)$, $\mu_\omega(s)$ and $\beta_\omega(s)$ are adjusted to $Q(s,\omega,a)$, $\mu(s,\omega)$ and $\beta(s,\omega)$ respectively. 

We first modify the loss function of $Q(s,\omega,a)$. The idea of one-step temporal-difference is still effective, only if the probability of termination with respect to the current option $\omega_t$ and the next state $s_{t+1}$ is considered. Specifically, $U(\omega_t,s_{t+1})$ is introduced to compute one-step estimated values. If the termination does not happen, the original estimated value $Q(s_{t+1},\omega_t,a_t)$ can be directly applied. However, if terminating, the greedy approach is employed to estimate through the maximum of all options, i.e.,
\begin{gather}
\begin{split}
U(\omega_t,s_{t+1})&=(1-\beta(s_{t+1},\omega_t))Q(s_{t+1},\omega_t,a_t)\\
&+\beta(s_{t+1},\omega_t)\mathop{max}\limits_{\bar{\omega}}Q(s_{t+1},\bar{\omega},a_{t+1}).
\end{split}
\end{gather}
As a result, $L_{critic}$ could still be represented by the mean square error of modified $\delta_t$:
\begin{equation}
\begin{split}
\delta_t=r+\gamma U&(\omega_t,s_{t+1})-Q (s_t,\omega_t,a_t), \\
L_{critic}&=E_{s_t\sim\rho^\mu}\delta_t^2.
\label{eq:op_critic}
\end{split}
\end{equation}

When it comes to $L_{fu}$, analogous to Equation~\ref{eq:actor}, deterministic form of Intra-option (i.e., low-level) policy gradient with respect to the extended state $(s,\omega)$ could be written as:
\begin{equation}
	\nabla _{\theta} L_{fu}=-E_{(s_t,\omega_t)\sim \rho ^{\mu}}[\nabla_\theta \mu_\theta(s_t,\omega_t) \nabla_a Q(s_t,\omega_t,a_t)],
\end{equation}
where the gradient of $Q$ with respect to $a$ should be computed in the case of $a_t=\mu_\theta (s_t,\omega_t)$.

Finally, we follow the \textit{Termination Gradient Theorem}~\cite{bacon2017option} to formulate the loss of the termination function, i.e.,
\begin{align*}
	\begin{split}
	A(s_{t+1}, \omega_t)=Q(&s_{t+1},\omega_t,a_{t+1})-	\mathop{max}\limits_{\bar\omega}Q(s_{t+1},\bar\omega,a_{t+1}), \\
	L_{term}&=\beta(s_{t+1},\omega_t)A(s_{t+1}, \omega_t),
	\end{split}
\end{align*}
where the one-step evaluation $Q(s_{t+1},\omega_t,a_{t+1})$ from the critic indicates an update margin of the termination function, with the baseline reduction technique~\cite{schulman2015high} applied for stability.

\subsection{Training Strategy}
While the final $L_{critic}$ in the TC3 and TC3-Options are directly described in the Equation~\ref{eq:critic} and~\ref{eq:op_critic} respectively, the $L_{actor}$ in both of our models are composed of different parts.

Considering another obvious prior that the funding progress of a campaign increases monotonously, we add the following restriction:
\begin{equation}
L_{reg}=\sum^T_{t=2}(\hat{p}_t-\hat{p}_{t-1})^2I_{[\hat{p}_t <\hat{p}_{t-1}]},
\end{equation}
where the square errors of adjacent predicted funding progress are penalized only when $\hat{p}_{t-1}$ is greater than $\hat{p}_t$.

As a result, when it comes to $L_{actor}$ in the basic TC3, the losses of future estimates, i.e., $L_{fu}$ and past experiences, i.e., $L_{pa}$ are combined, along with $L_{reg}$:
\begin{equation}
L_{actor} = L_{fu} + \lambda_1 L_{pa} + \lambda_2 L_{reg}.
\end{equation}
While $L_{actor}$ in the TC3-Options still needs to integrate the loss of termination function, i.e., $L_{term}$:
\begin{equation}
L_{actor} = L_{fu} + \lambda_1 L_{pa} + \lambda_2 L_{reg} + \lambda_3 L_{term}.
\end{equation}

Here, we do not prescribe how to acquire high-level policy $pi$ since many approaches could be utilized such as primitive policy gradient, planning or temporal difference updates. However, computing $Q(s,\omega,a)$ in addition to $\pi(\omega|s)$ seems to be wasteful. Therefore, we obtain $\pi$ according to $Q(s,\omega,a)$, with adaptive epsilon greedy policy adopted to keep the balance between exploration and exploitation.

Some other tricks may need to be explained. The first one is copying the actor and critic network to the target ones. While the update of parameters in the target networks is delayed~\cite{lillicrap2015continuous}. Secondly, we store complete trajectories in the experience replay (i.e., $H_{1:T}=s_1,a_1,r_1,s_2,a_2,r_2,...,s_T,a_T,r_T$) instead of one-step interactions (i.e., $H_{t:t+1}=s_t,a_t,r_t,s_{t+1}$)~\cite{heess2015memory}. The common purposes are for learning more stably in addition to accelerating the convergence.

\section{Experiments}
In this section, we first introduce the dataset we collect from \textit{Indiegogo}. Then, the detailed experimental setup follows. Finally, the results of experiments are demonstrated, especially the validation of the U-shaped pattern.

\subsection{Dataset Description}

We collect a real-world dataset from Indiegogo, which is a famous reward-based crowdfunding platform. The dataset includes 14,143 launched campaigns from July 2011 to May 2016, soliciting over 18 billion funds from 217,156 backers. In addition, there are totally 98,923 perks and 240,922 comments, along with 1,862,097 backing records. According to the statistics, in our dataset, 62.54\% of campaigns have pledged funding duration between 30 and 60 days. However, there are still 7.14\% of campaigns whose funding duration is between 15 and 25 days.
\begin{table}[h]
	\small
	\caption{The information of features}
	\label{tab:data}
	\begin{tabular}{p{1.2cm}<{\centering}|p{4.0cm}<{\centering} |p{1.6cm}<{\centering}}
		\hline
		\textbf{Level} & \textbf{Features} & \textbf{Type} \\
		\hline
		\multirow{9}{*}{Static} & campaign description & textual \\
		& perk description & textual \\
		& campaign's category & categorical \\
		& creator's type & categorical \\
		& funding duration & numerical \\
		& pledged goal & numerical \\
		& number of perks & numerical \\
		& number of comments & numerical \\
		& max/min/avg price of perks & numerical \\
		\hline
		\multirow{4}{*}{Dynamic} & comments & textual \\
		& number of day started & numerical \\
		& number of day left & numerical \\
		& current schedule & numerical \\
		\hline
	\end{tabular}
\end{table}

\subsection{Experimental Setup}
\noindent \textbf{Parameter Setting.}
For the static features of campaigns, we adopt one-hot encoding for categorical features and word2vec embedding~\cite{mikolov2013distributed} for textual features (each with a 50-dimensional vector). Finally, all kinds of static features are concatenated to 182-dimensional vectors. While for the dynamic features, they are 19-dimensional vectors composed of textual comments (16-dimensional by word2vec embedding), day information (2-dimensional) and current funding progress. Specially, all kinds of features are scaled by Min-Max normalization.

Additionally, considering the shortest funding duration of campaigns in our training data and testing data is 15, we set the length of days to be predicted to be 6, 7, 8, 9, 10 respectively. With respect to the coefficients of regularization terms, we set $\lambda_1$, $\lambda_2$ and $\lambda_3$ to be 100, 1 and 1 respectively.

\noindent \textbf{Evaluation Metrics.}
First we randomly select 10\% of all campaigns in our dataset as the testing set. Then, considering the task is to predict the series of funding progress for campaigns in the future, we adopt the following three metrics to evaluate the performance, i.e., root mean square error (RMSE), mean absolute error (MAE) and mean absolute percentage error (MAPE). 
Specifically, for campaign $i$ with pledged duration of $T$ days, given its series of real funding process in the final $\tau$ days (i.e., $p_{T-\tau+1}^\textit{i},p_{T-\tau+2}^\textit{i},...,p_{T}^\textit{i}$) and the series of predicted funding process (i.e., $\hat{p}_{T-\tau+1}^\textit{i},\hat{p}_{T-\tau+2}^\textit{i},...,\hat{p}_{T}^\textit{i}$), the performance could be measured by:

\begin{align}
	MAE&=\frac{1}{I*\tau}\sum_{i=1}^I\sum_{k=1}^\tau\left |p^i_{T-\tau+k}-\hat{p}^i_{T-\tau+k}\right |, \\
	RMSE&=\sqrt{\frac{1}{I*\tau}\sum_{i=1}^I\sum_{k=1}^\tau(p^i_{T-\tau+k}-\hat{p}^i_{T-\tau+k})^2}, \\
	MAPE&=\frac{100\%}{I*\tau}\sum_{i=1}^I\sum_{k=1}^\tau \left |\frac{p^i_{T-\tau+k}-\hat{p}^i_{T-\tau+k}}{p^i_{T-\tau+k}}\right|.
\end{align}

\noindent \textbf{Benchmark Methods.}
\begin{itemize}
	\item \textbf{VAR} \textit{(Vector Autoregression)}~\cite{sims1980macroeconomics} models generalize the univariate autoregressive (AR) model by allowing for more than one evolving variable.
	\item \textbf{RFR} \textit{(Random Forest Regression)} is one of the ensemble methods that could balance different regression results of all decision trees.
	\item \textbf{SWR} \textit{(Switching Regression)}~\cite{zhao2017tracking} is a variant of regression model that combines campaign-level and perk-level regression results.
	\item \textbf{MLP} \textit{Multi-layer Perceptron}~\cite{bengio2009learning} is a kind of artificial neural network that performs well in dealing with high-dimensional features.
	\item \textbf{SMP-A}~\cite{jin2019estimating} is a variant of Seq2Seq model that using an encoder to track the history dynamics and a decoder to predict the future dynamics, along with the monotonously increasing prior.
	\item \textbf{TC3} is our proposed basic model to utilize actor-critic architecture to simulate decision-making process between investors and campaigns.
	\item \textbf{TC3-Options} is the complete model that combine basic TC3 with a structure of options to utilize the U-shaped pattern in crowdfunding.
\end{itemize}

\begin{table*}[pt]
	\small
	\centering
	\caption{Performance on Funding Progress Prediction.}
	\begin{tabular}{c|c|ccccccc}
		\toprule
		Test Length & Metric & VAR   & RFR    & SWR   & MLP   & SMP-A & TC3   & TC3-Options \\
		\midrule
		\midrule
		\multirow{3}[2]{*}{6-day} & MAE   & 0.1245  & 0.1392  &    0.1104   & 0.0648  & 0.0372  & 0.0234  & \textbf{0.0201 } \\
		& RMSE  & 0.1935  & 0.2115  &   0.1695    & 0.1368  & 0.0927  & 0.0681  & \textbf{0.0435 } \\
		& MAPE  & 41.23\% & 40.23\% &    35.03\%   & 22.54\% & 12.82\% & 11.72\% & \textbf{9.05\%} \\
		\midrule
		\multirow{3}[2]{*}{7-day} & MAE   & 0.1668  & 0.1533  &   0.1365    & 0.0990  & 0.0399  & 0.0258  & \textbf{0.0237 } \\
		& RMSE  & 0.2790  & 0.2535  &    0.2046   & 0.1671  & 0.0963  & 0.0771  & \textbf{0.0495 } \\
		& MAPE  & 52.65\% & 46.15\% &   36.21\%    & 26.97\% & 13.66\% & 12.27\% & \textbf{9.27\%} \\
		\midrule
		\multirow{3}[2]{*}{8-day} & MAE   & 0.1515  & 0.1443  &    0.1317   & 0.0921  & 0.0420  & 0.0267  & \textbf{0.0252 } \\
		& RMSE  & 0.2574  & 0.2401  &   0.1989    & 0.1542  & 0.1080  & 0.0792  & \textbf{0.0504 } \\
		& MAPE  & 47.84\% & 43.63\% &   35.65\%    & 23.82\% & 12.58\% & 11.26\% & \textbf{8.76\%} \\
		\midrule
		\multirow{3}[2]{*}{9-day} & MAE   & 0.1494  & 0.1437  &   0.1257    & 0.0960  & 0.0414  & 0.0258  & \textbf{0.0249 } \\
		& RMSE  & 0.2466  & 0.2268  &   0.1944    & 0.1581  & 0.0942  & 0.0759  & \textbf{0.0483 } \\
		& MAPE  & 44.84\% & 40.96\% &   34.15\%    & 24.69\% & 11.91\% & 10.37\% & \textbf{8.18\%} \\
		\midrule
		\multirow{3}[2]{*}{10-day} & MAE   & 0.1467  & 0.1338  &  0.1209    & 0.0987  & 0.0408  & 0.0249  & \textbf{0.0237 } \\
		& RMSE  & 0.2394  & 0.2197  &    0.1905   & 0.1647  & 0.1011  & 0.0741  & \textbf{0.0465 } \\
		& MAPE  & 42.38\% & 38.73\% &   32.20\%    & 25.91\% & 11.31\% & 9.71\% & \textbf{7.60\%} \\
		\bottomrule
	\end{tabular}%
	\label{tab:main}%
\end{table*}%

%
\begin{figure}[t]
	\centering
	\includegraphics[width=0.98\columnwidth]{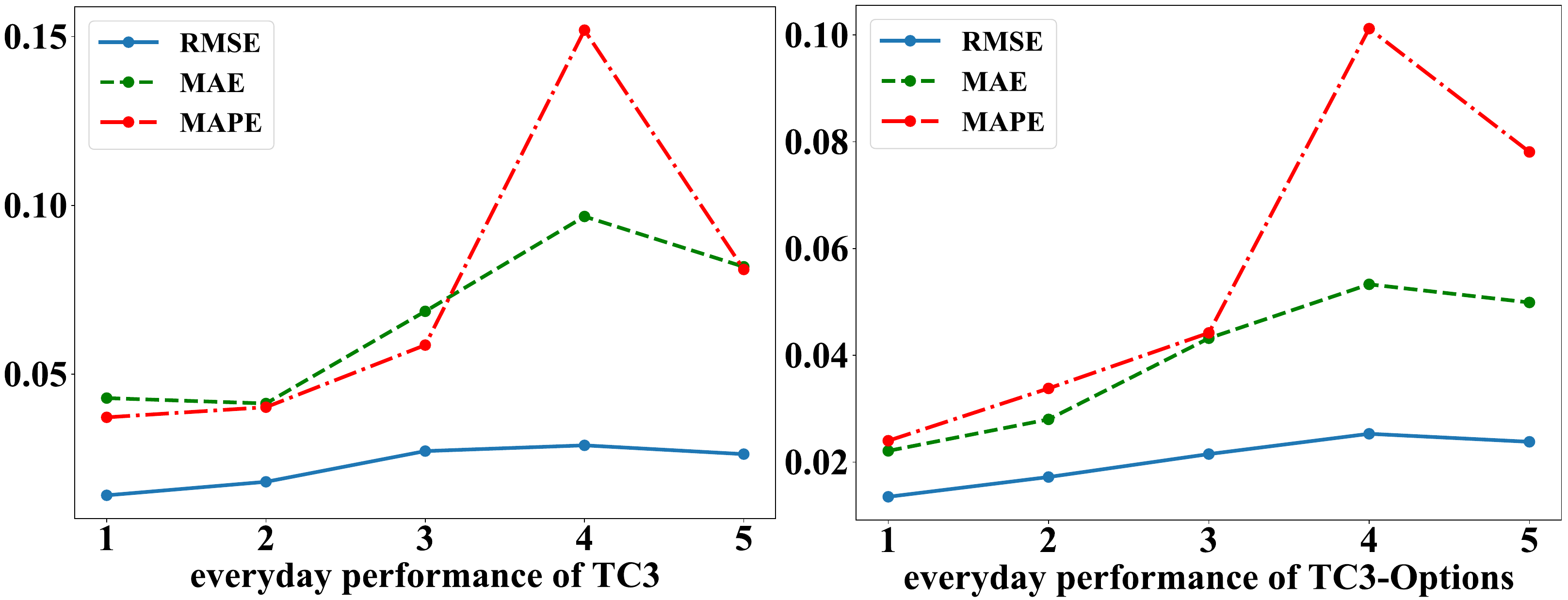}
	\caption{The first 5 days performance of TC3 (left) and TC3-Options (right) (Test length = 10).}
	\label{fig:ed_error}
\end{figure}

\subsection{Experimental Results}
\subsubsection{Performance on Funding Progress Prediction.}
Here, we demonstrate the performance comparisons on the funding progress prediction task. Table~\ref{tab:main} provides the results on RMSE, MAE, MAPE metrics, testing through last 6, 7, 8, 9, 10 days respectively. Overall, it could be observed that both of our proposed models (TC3, TC3-Options) outperform the other baselines in all cases, which indicates that modeling the decision process between investors and campaigns might be helpful when tracking the dynamics in crowdfunding, especially when future tendencies are specially considered. Secondly, compared with TC3, TC3-Options performs better, which suggests that utilizing a well-learned pattern would improve the accuracy of prediction. However, improvements between TC3 and TC3-Options are more evident in RMSE and MAPE metrics instead of MAE metrics. It is possible that the RMSE metric decreases because of the smoother distribution of errors while the MAPE metric falls due to the more accurate prediction when it comes to campaigns with fewer contributions. Thirdly, neural network models (MLP, SMP-A, TC3, TC3-Options) outperform the regression-based models (VAR, RFR, SWR) in a whole, which confirms that this kind of methods could better deal with high-dimensional features.

Furthermore, as the length of test days to be longer, the metrics do not show the monotonous increasing tendency, which does not agree with intuition. To further explore this, we specially measure the everyday performance of our proposed TC3 and TC3-Options (the number of option is 2) from the 1st to the 5th day when the test length is 10 days. The results are shown in the Figure~\ref{fig:ed_error}, which demonstrates that the error of the first and second day are smaller than the other days evidently. A likely explanation may be that since the length of test days is over one week, most campaigns are going through the gentle raise phrase at the beginning of the test period. As a result, smaller fluctuations seem to make it easier for the algorithm to predict. Actually, the MAPE of the first 5 days is 5.62\%, compared with the following 5 days of which the value is 9.54\%.

\begin{table}[h]
	\small
	\centering
	\caption{Influence of \#Options.}
	\begin{tabular}{c|ccccc}
		\toprule
		\#options & 1     & 2     & 3     & 4     & 5 \\
		\midrule
		\midrule
		MAE   & 0.0234  & 0.0201  & 0.0237  & 0.0246  & 0.0261  \\
		RMSE  & 0.0681  & 0.0435  & 0.0468  & 0.0531  & 0.0582  \\
		MAPE  & 11.72\% & 9.05\% & 10.40\% & 11.52\% & 11.06\% \\
		\bottomrule
	\end{tabular}%
	\label{tab:noption}%
\end{table}%

\subsubsection{Parameters Effects.}
In this subsection, we conduct a group of experiments to test the influence of numbers of options, where other parameters (length of test day, learning rate, training steps et al.) are kept the same. It should be reminded that when the number of option is 1, the TC3-Options model degenerates to the basic TC3 model. The results are shown in Table~\ref{tab:noption}. It is obvious that our model performs best when the number of options is 2. However, the model does not learn better when the number increases, despite it still outperforms the basic TC3 model. A likely explanation is that the model is forced to learn more sub-patterns hidden behind the data, while the data would not be complicated enough for so many patterns, hence, the model could be confused to select the proper option.

\begin{figure}
	\centering
	\subcaptionbox{\label{fig:shape}}
	{
		\includegraphics[width=0.445\columnwidth]{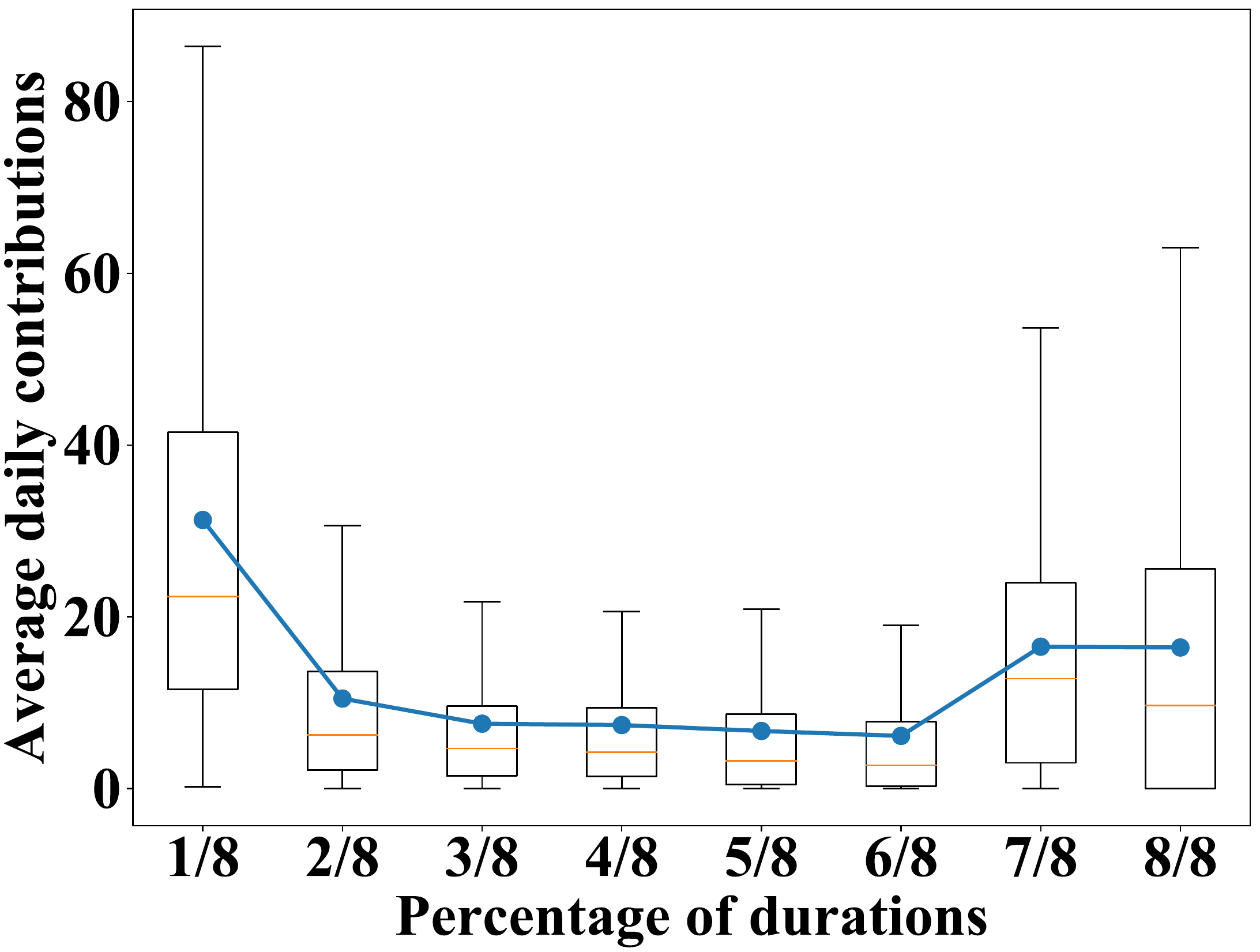}
	}
	\subcaptionbox{\label{fig:term}}
	{
		\includegraphics[width=0.50\columnwidth]{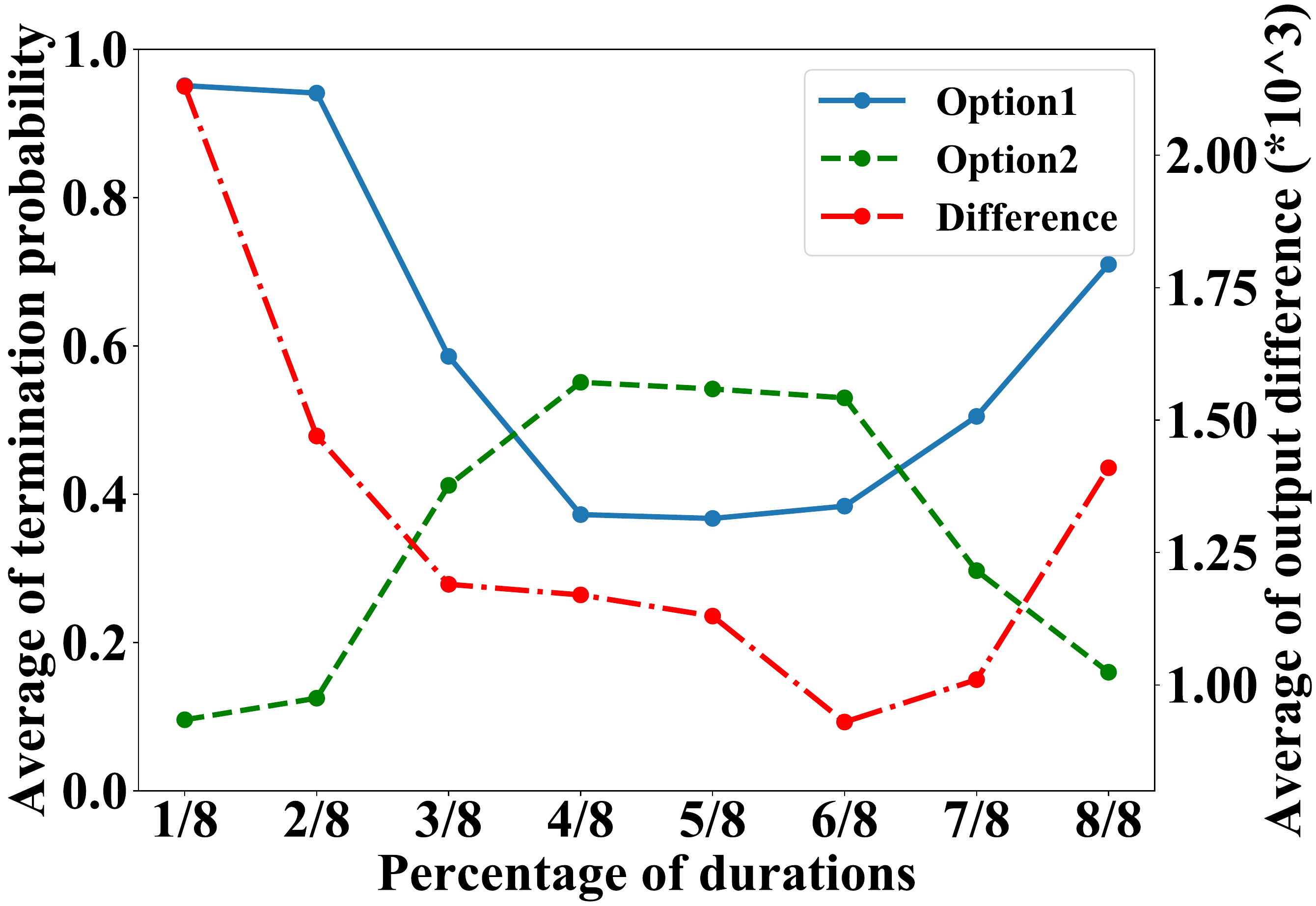}		
	}
	\caption{(a) Backing distribution from Dataset. (b) Termination value learned.}
	\label{fig:verify}
\end{figure}

\subsubsection{U-shaped Verification.}
Before we validate the learned U-shaped pattern, we conduct a statistical experiment to show the U-shaped pattern hidden behind the dataset. Concretely, we divide the whole funding cycle into eight parts according to the trajectory length and then calculate the mean percentage of the contributions. The results are shown on the Figure~\ref{fig:shape}, in which U-shaped pattern is obvious.

Finally, we verify whether or not the pattern learned by our model is the U-shaped pattern when the number of options is 2. Ideally, a well-trained high-level policy could select proper options according to input states, in addition to the termination function could instruct the appropriate probability to terminate in the current option. Significantly, it is not in every step that the high-level policy would select the option. Hence, a more effective approach is to observe the values of the termination function in every step. Owing to the experimental setting, the options would terminate with the probability of $1-\beta(s,\omega)$. If the termination value of the current option is low, the model is more likely to terminate and select a new option in the next day.

The average values of $\beta(s,\omega_1)$ and $\beta(s,\omega_2)$ in different periods are shown on the Figure~\ref{fig:term}. The partition of the whole funding cycle is the same as above. Obviously, the results illustrate that the first option has high termination probability in the start and end stages while the second option shows relatively high termination value in the middle phase, which implies that the low-level policy $\mu$ of the first option learns from the fast-growing sub-patterns while the policy of another option learns from the slow-growing sub-patterns. This could also be proved by the average difference of outputs between the two sub-policies described in the other y-axis, which discloses that mean outputs from $\mu$ of Option 1 are all greater than those from $\mu$ of Option 2.

\section{Conclusions}
In this paper, we presented a focused study on forecasting dynamics in crowdfunding with an exploratory insight. Inspired by techniques of reinforcement learning, especially hierarchical reinforcement learning, we first propose a basic model which could forecast the funding progress based on the decision-making process between investors and campaigns. Then, through observing the typical U-shaped pattern behind the crowdfunding series, we design a specific actor component with a structure of options to fit for various sub-patterns in different stages of funding cycles. As a result, we validated the effectiveness of our proposed TC3 and TC3-Options models by comparing with other state-of-the-art methods. Moreover, extra experiments are conducted to demonstrate the entire pattern learned by TC3-Options is exactly the U-shaped one.

In the future, we will generalize this framework to capture dynamic pattern-switching process in other tasks that could be modeled as sequential decision-making processes.

\section*{Acknowledgements}
This research was partially supported by grants from the National Key Research and Development Program of China (No. 2016YFB1000904), the National Natural Science Foundation of China (Grants No. 61672483, 61922073), and the Science Foundation of Ministry of Education of China \& China Mobile (No. MCM20170507). Qi Liu acknowledges the support of the Young Elite Scientist Sponsorship Program of CAST and the Youth Innovation Promotion Association of CAS (No. 2014299).

\small
\bibliographystyle{aaai}
\bibliography{2711-References}

\end{document}